\documentclass[fleqn,10pt]{wlscirep}
\usepackage[utf8]{inputenc}
\usepackage[T1]{fontenc}
\usepackage{multirow}
\usepackage{hyperref}
\usepackage{graphicx}

\title{A framework for recipe data structure with applications for culinary and nutritional insights}

\author[1,4,5]{Mansi Goel}
\author[2]{Sumit Bhagat}
\author[2]{Saloni Srivastava}
\author[1]{Malav Patel}
\author[2]{Hardi Parikh}
\author[3]{Shlok Vinodkumar Mehroliya}
\author[1,4,5,$\ast$]{Ganesh Bagler}
\affil[1]{Department of Computational Biology, Indraprastha Institute of Information Technology Delhi (IIIT-Delhi), New Delhi, 110020, India}
\affil[2]{Department of Computer Science, Indraprastha Institute of Information Technology Delhi (IIIT-Delhi), New Delhi, 110020, India}
\affil[3]{Department of Social Sciences and Humanities, Indraprastha Institute of Information Technology Delhi (IIIT-Delhi), New Delhi, 110020, India}
\affil[4]{Infosys Centre for Artificial Intelligence, Indraprastha Institute of Information Technology Delhi (IIIT-Delhi), New Delhi, 110020, India}
\affil[5]{Center of Excellence in Healthcare, Indraprastha Institute of Information Technology Delhi (IIIT-Delhi), New Delhi, 110020, India}

\affil[*]{Corresponding author: bagler@iiitd.ac.in}

\begin{abstract}
Cooking is a complex process that transforms raw ingredients into delicious and nutritious dishes, yet the recipes that encode this process remain largely free text; readable by people but not directly computable. Existing recipe collections capture fragments of this information, but no shared representation links a recipe's structured ingredient composition, its geo-cultural provenance, and its nutritional profile within a single queryable schema. We address this representation gap by formalizing a framework for recipe data structure that decomposes each recipe into typed ingredient entities, grounds those entities in a reference nutritional database, and annotates them with geo-cultural and dietary context. We present RecipeDB2, a structured compilation of 128,942 recipes with 35,474 ingredients from 32 regions and 99 countries. Ingredient phrases are parsed into seven culinary attributes using a transformer-based named-entity model; ingredients are linked to the USDA reference tables through a BERT embedding strategy (F1 = 87.90 on a manually adjudicated set of the 200 most frequent ingredients), yielding 148 nutritional parameters per mapped ingredient; a Random Forest classifier propagates 34 ingredient categories across the full vocabulary; and a deterministic, conservative rule set assigns each recipe a dietary style. Through RecipeDB2~(\href{https://cosylab.iiitd.edu.in/recipedb2/}{https://cosylab.iiitd.edu.in/recipedb2/}), we demonstrate a scalable framework for making recipes computable, turning culinary heritage (long treated as an artistic rather than a quantitative object) into a data-driven analysis. 
\end{abstract}

\begin{document}

\flushbottom
\maketitle

\thispagestyle{empty}

\section*{Introduction}

Cooking is the art of transforming raw ingredients into flavorful and nutritious dishes---a skill passed down through generations via recipes~\cite{crosby2020cook, pollan2014cooked}. These recipes encode essential knowledge about processing natural ingredients, making them palatable and nutritionally beneficial. Over time, distinct cooking styles, referred to as cuisines, have emerged, conforming to geo-cultural constraints. Despite technological advancements and evolving eating habits, daily dietary intake remains deeply rooted in cultural traditions~\cite{gatley2016significance}. Cooking has been pivotal in the evolution of large brain sizes in \emph{Homo sapiens}~\cite{wrangham2009catching} and plays a crucial role in shaping the gut microbiome~\cite{carmody2019cooking}. Thus, recipes serve as a bridge between our taste preferences and health outcomes.

As part of the Computational Gastronomy paradigm, building a structured and comprehensive repository of annotated recipes is essential to decode the complex interplay between taste, nutrition, and health in recipes~\cite{bagler2024computational, goel2022computational}. Traditionally, recipes have been passed down orally and, more recently, as written records~\cite{wharton2010recipes}. However, these formats remain unstructured and unannotated, posing challenges for computational analysis. In recent years, various datasets have been developed in the context of recipes such as Recipe1M~\cite{marin2021recipe1m+}, Epicurious~(\href{https://www.epicurious.com/}{https://www.epicurious.com/}), Yummly~(\href{https://www.yummly.com/}{https://www.yummly.com/}), and RecipeNLG~\cite{bien2020recipenlg}. Recipe1M is a large dataset containing over one million recipes sourced from various online platforms. However, its unstructured format comprising recipes, ingredients, and images often lacks critical metadata, including cooking techniques, preparation methods, and nutritional information. The Epicurious dataset offers a more structured format with well-defined ingredient lists and cooking instructions. It includes user ratings, preparation times, and seasonal categories but lacks nutritional data. Similarly, Yummly serves as a recipe discovery platform with a reasonably structured dataset that features ingredients, cooking steps, and dietary tags, yet it does not encompass comprehensive nutritional information. RecipeNLG~\cite{bien2020recipenlg}, designed primarily for natural language generation tasks, provides structured recipes but lacks a broader nutritional or cultural context. 

Other widely used datasets have gained prominence in recent studies. The Food.com dataset contains over 400,000 user-contributed recipes along with reviews, ratings, and user interactions, making it a valuable resource for studying user preferences and recommendation systems. Similarly, MealRec is a large-scale dataset designed to support food recommendation and meal planning, comprising over one million meal sessions annotated with contextual features such as time, user behavior, and co-consumed recipes. These datasets highlight the growing diversity and reinforce the need for structured, annotated repositories to advance research in this interdisciplinary field. RecipeDB~\cite{batra2020recipedb} is a structured dataset that overcomes most of these shortcomings, but has a limited size and low mapping efficiency between ingredients and their nutritional profile. The Named Entity Recognition (NER) performance in RecipeDB was low, which affected the prediction of named entities and the further mapping to their nutritional correlates. 

We address these limitations with a framework for recipe data structure, RecipeDB2. Our contributions are threefold. First, we define a compositional schema in which a recipe is represented as a set of typed ingredient entities, each grounded in a reference nutritional database and enriched with geo-cultural and dietary annotations. Second, we instantiate this schema at scale by combining an improved transformer-based NER model, an embedding-based ingredient-to-USDA mapping, a supervised ingredient-category classifier, and a deterministic dietary-style rule set, validating each component against manually curated ground truth. Third, because these components form a pipeline, we characterize how the reliability of each stage bounds the layers that depend on it, so that the resulting estimates are interpretable rather than opaque. RecipeDB2 builds on RecipeDB~\cite{batra2020recipedb} while substantially extending both its size and the richness of its annotations, such as geo-cultural cuisine, dietary style, cooking techniques, utensils, ingredient detail, and nutritional profiles. This repository makes recipes computable, opening up new avenues for data-driven exploration of global cuisines by accounting for their culinary nuances.

Among other data resources related to the objectives of RecipeDB, FooDB focuses on the compilation of food chemicals (http://foodb.ca). FoodBase~\cite{popovski2019foodbase} provides an annotated food entity resource and has been extended to include corpora such as CafeteriaSA and CafeteriaFCD, supporting sentiment analysis and food composition analysis. Furthermore, recent developments in the domain leverage NLP workflows to extract relations between food and biomedical entities---examples include FoodChem~\cite{cenikj2021foodchem} and FooDis~\cite{cenikj2023foodis}, which facilitate the construction of domain-specific food knowledge graphs. Notably, Cenikj et al.~\cite{cenikj2023language} demonstrate the integration of large language models to construct large-scale food and biomedical knowledge graphs, illustrating the potential of advanced NLP methods in food informatics. Other knowledge graph efforts, such as FoodKG\cite{haussmann2019foodkg}, further exemplify semantics-driven approaches for food recommendation and reasoning. Some other databases, such as FlavorDB~\cite{garg2018flavordb}, FlavorDB2~\cite{goel2022flavordb2}, BitterDB~\cite{wiener2012bitterdb}, and SuperSweet~\cite{ahmed2010supersweet}, focused on taste and olfaction, attempting to address the interaction of natural entities with human sensory machinery. Databases such as NutriChem~\cite{jensen2015nutrichem}(nutritional factors), and DietRx emphasized the food-nutrition-health axis. 

\begin{figure*}[!htb]
    \includegraphics[width=\textwidth, angle=0]{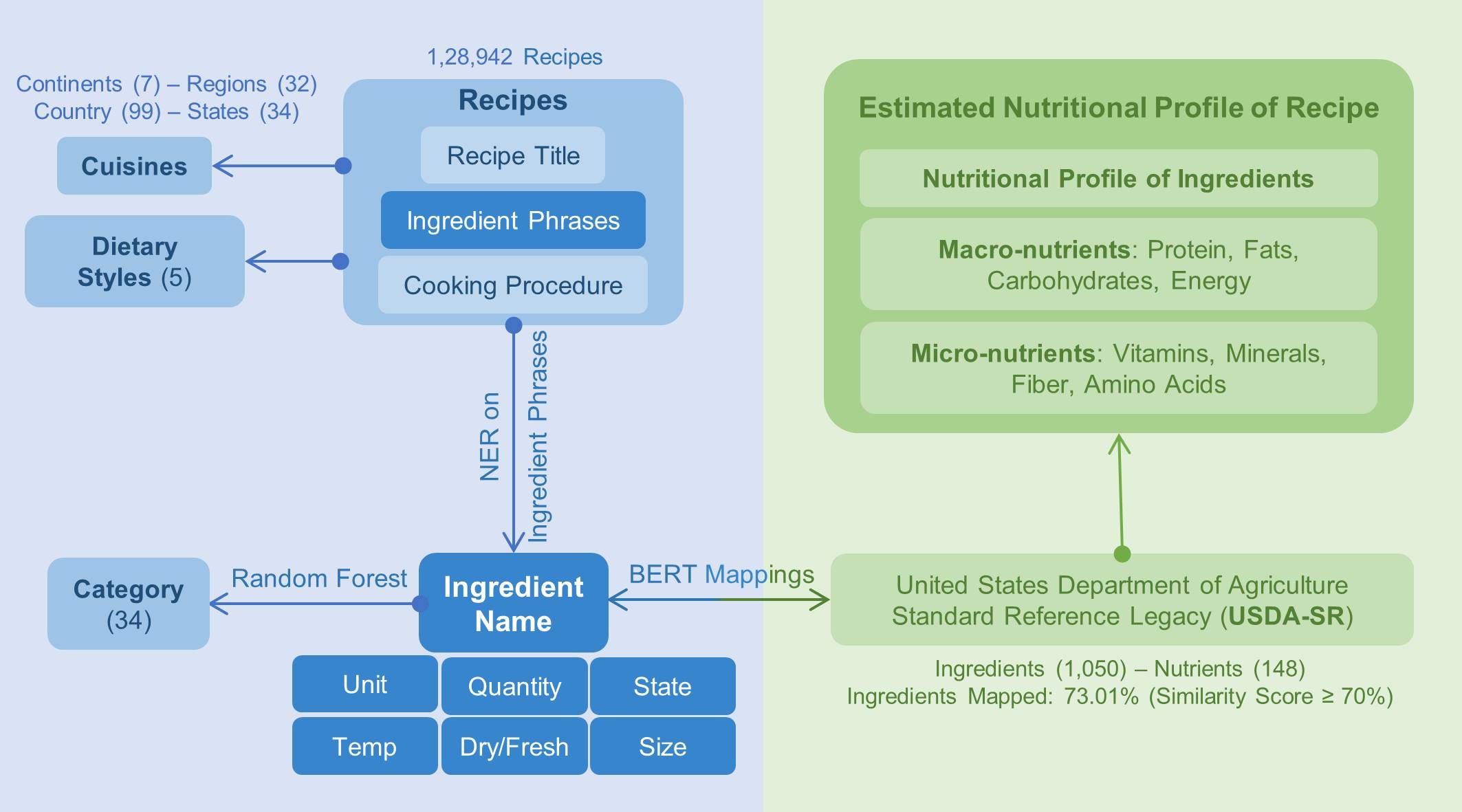}
  \caption{The flow diagram depicting the framework for recipe data structure involving data compilation, curation, annotations, mapping, and estimation of the nutritional profiles. Building this framework involved the application of multiple machine-learning models. A deep-learning based named entity recognition model was used for extracting relevant culinary elements from the ingredient phrases. BERT was implemented to map ingredients from the recipes to those in the USDA nutrition table. A Random Forest model was used to predict the ingredient category. This framework for recipe data structure is generic and scalable.}
  \label{pipeline}
\end{figure*} 

RecipeDB2 is an expanded and annotated resource designed to probe the relationships between culinary practices and nutrition (Figure~\ref{pipeline}). Recipes are broken down into their culinary elements using advanced algorithms, creating a searchable database that reflects geo-cultural contexts, dietary preferences (Vegan, Pescetarian, Lacto-Vegetarian, Ovo-vegetarian,  Ovo-Lacto vegetarian), cooking methods, and ingredient attributes. It provides a detailed breakdown of each ingredient, including name, quantity, unit, state, and additional characteristics. RecipeDB2 also integrates data from the USDA database, offering deeper nutritional insights with consequences for health. The recipe framework for recipe data structure presents RecipeDB2 as an illustration. This updated resource empowers researchers, nutritionists, chefs, and food enthusiasts to explore the complex relationships between food, culture, nutrition, and sensory experiences.

\section*{Framework Overview}
\subsection*{Data compilation}
We created an extensive, structured database of recipes by appending the RecipeDB~\cite{batra2020recipedb} (118,171 recipes) with those obtained from Archana's Kitchen (\href{https://www.archanaskitchen.com/}{https://www.archanaskitchen.com/}) (9,730 recipes) and Awesome Cuisine ~(\href{https://www.awesomecuisine.com/}{https://www.awesomecuisine.com/}) (1,132 recipes). These datasets were selected based on several critical criteria, including their structural uniformity, the availability of geo-cultural mapping for ingredients and dishes, and the substantial number of recipes. The dataset encompasses a wide variety of information about each recipe, including the recipe title, list of ingredients, servings, detailed cooking instructions, preparation time, cooking time, total time, and associated recipe images.

To facilitate detailed data extraction, we divided each recipe into two main parts: the ingredients section and the cooking instructions section. This division helps to isolate and analyze relevant components more effectively. Our goal was to extract structured elements and relevant culinary information from these sections, such as ingredient names, quantities, units, and preparation steps, for further analysis. This structured data has applications in tasks ranging from nutritional estimation and recipe generation to the geo-cultural classification of dishes.

\subsection*{Geo-cultural mapping of recipes}
Every recipe in the dataset was mapped to its geo-cultural correlate at different levels of hierarchy: Continent, Region, Sub-region (Country), Sub-sub-region (State, where applicable), and Sub-sub-sub-region (City, where applicable). This multi-level mapping provides a deeper understanding of each recipe's geographical and cultural context. The continent and country-level mapping of recipes was obtained from their original data sources. For the region-level mapping, we employed a more flexible approach, focusing on culinary and cultural similarities rather than strictly adhering to geo-political boundaries. With such an approach, one can capture shared culinary traditions and influence across political borders.

Countries were grouped into regions by virtue of their shared geo-cultural correlates, and regions were further attached with coarse-grained labels of continents. The sub-region, sub-sub-region, and sub-sub-sub-region levels provide a more granular view down to specific countries, states, or cities, where available, further refining the geo-cultural context of the recipes. The data comprised 7 continents, 32 regions, 99 sub-regions, 34 sub-sub-regions, and 12 sub-sub-sub-regions. Supplementary Table S1 presents the schema of the continent and region mapping, providing clarity to geo-cultural mappings of recipes.

\subsection*{NER on ingredient section}
We extracted `named entities' from the ingredients section of each recipe to capture structured data essential for nutritional analysis. We used the state-of-the-art spaCy-transformer model~\cite{goel2024deep} to extract the named entities~\cite{diwan2020named}. Through this process, we identified seven critical attributes for providing relevant nutritional information about a recipe. These attributes allow us to standardize and quantify ingredients for further analysis. The attributes identified are as follows:

\begin{itemize}
    \item Name: The ingredient's name, such as salt or pepper, provides the recipe's foundational component.
    \item Unit: The unit of measurement associated with the ingredient, such as gram, cup, tablespoon, or teaspoon, specifying how the ingredient's quantity is measured.
    \item Quantity: The specific amount of the ingredient, paired with the unit, for example, 1 teaspoon, 10 grams, or 2 cups, to provide a precise measurement for recipe scaling and nutritional estimation.
    \item State: The processing state or form of the ingredient, such as ground, chopped, thawed, or blanched, reflecting its preparation before use.
    \item Size: Any portion size descriptors for the ingredient, such as small, medium, or large, which further refine the quantity and help standardize serving sizes in recipes.
    \item Temperature: The temperature condition of the ingredient before or during preparation, such as hot, cold, or room temperature, can affect the cooking process and the ingredient's properties.
    \item Dry/Fresh: A classification that identifies whether the ingredient is dry (e.g., dried herbs, spices) or fresh (e.g., fresh basil, vegetables), which influences the flavor profile and shelf life of the ingredient.
\end{itemize}

\subsection*{Standard units conversion}
\begin{figure}[!htb]
\centering
    \includegraphics[width=0.6\textwidth, angle=0]{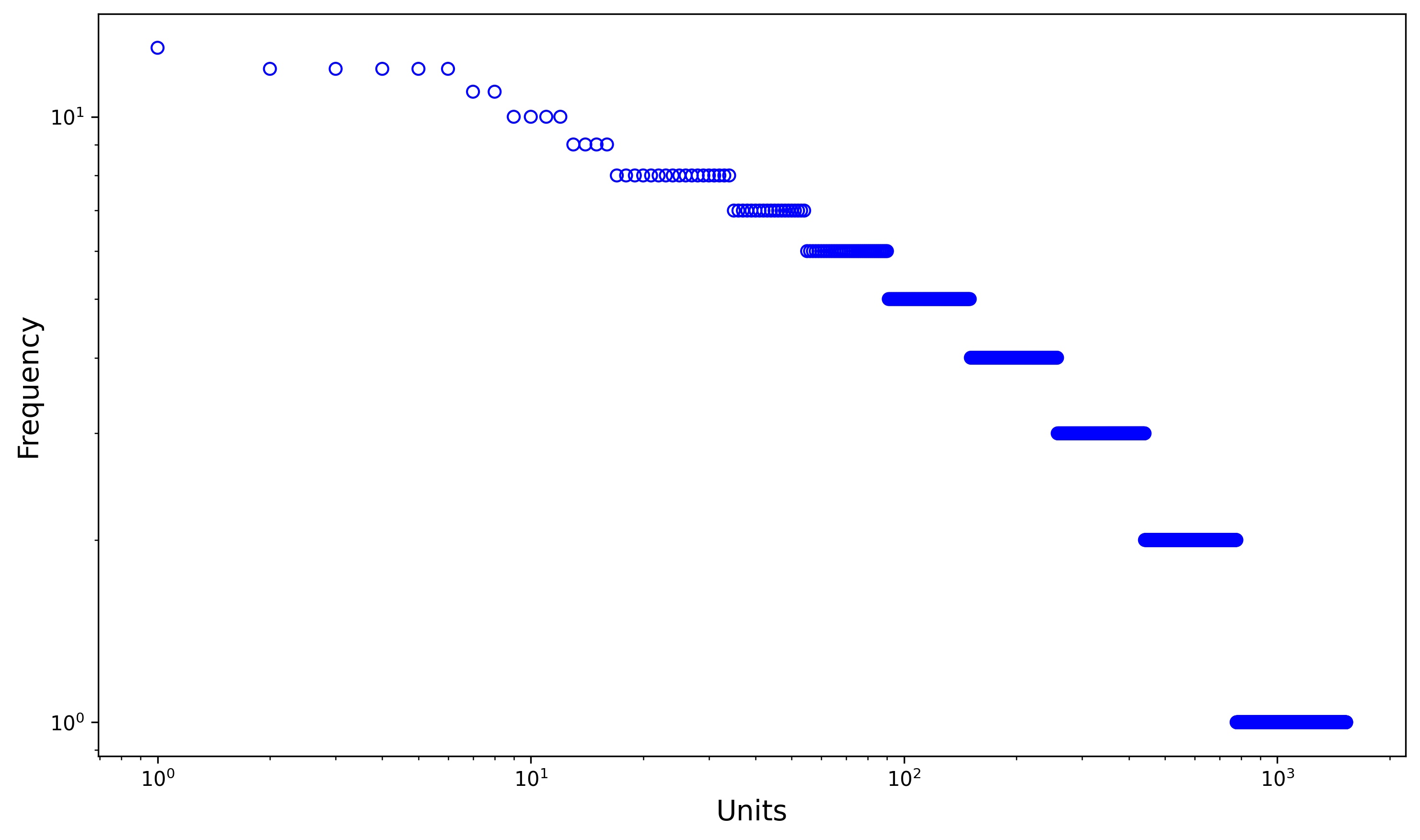}
  \caption {Frequency rank distribution of units. The data indicates a presence of a thick-tailed (scale-free) distribution suggesting that few units are present in disproportionately high frequency.}
  \label{unit_freq}
\end{figure}

Ingredients in recipes most often come with a variety of measurement units. For example, salt can be measured in tablespoons, teaspoons, pinches, dashes, packets, envelopes, etc. Similarly, other ingredients are paired with diverse units of measurement. RecipeDB2 data features an extensive collection of 35,474 unique ingredients and 1,163 distinct measurement units. Table~\ref{tab:popular_ing_unit} shows the 20 most popular ingredients and their unique units, where salt (53234; 41.28\%), garlic (42789; 33.18\%), onion (37195; 28.84\%), water (27799; 21.55\%), and butter (26624; 20.64\%) were among the most popular ingredients. On the measurement side, Figure~\ref{unit_freq} shows the frequency rank distribution of units, commonly used units in recipes are cup (291,756), teaspoon (224,404), tablespoon (187,824), lb (417,70), and ounce (39,571). This wide variation in ingredients and measurement units highlights the complexity of recipe creation and cooking.

\begin{table*}[]
\centering
\resizebox{\textwidth}{!}{%
\begin{tabular}{|c|c|c|c|}
\hline
Ingredients   & Frequency & Units & Unit Count \\ \hline
salt  & 53234 & {[}`cup', `dash', `g', `ml', `ounce', `pinch', `tablespoon', `teaspoon', `tsp'{]} & 9  \\ \hline
garlic & 42789 & {[}`bulb', `clove', `cup', `g', `halved',   `ounce', `pressed', `tablespoon', `teaspoon', 'tsp'{]}  & 10 \\ \hline
onion  & 37195 & {[}`cup', `g', `halved', `kg', `lb', `ounce', `pound', `slice', `tablespoon', `teaspoon'{]}   & 10 \\ \hline
water  & 27799 & {[}`can', `cup', `g', `gallon', `liter', `ml', `ounce', `pint', `quart', `tablespoon', `teaspoon'{]}   & 11 \\ \hline
butter & 26624 & {[}`cup', `g', `gram', `lb', `ml', `ounce', `pound', `stick', `tablespoon', `tbsp', `teaspoon', `tsp'{]}   & 12 \\ \hline
egg    & 22901 & {[}`cup', `dozen', `extra large', `large', `whole'{]}  & 5   \\ \hline
sugar  & 22144 & {[}`cup', `dash', `g', `gram', `kg', `lb', `ml', `ounce', `pinch', `tablespoon', `tbsp', `teaspoon', `tsp'{]} & 13 \\ \hline
olive oil & 21257 & {[}`cup', `dash', `ml', `ounce', `tablespoon', `tbsp', `teaspoon'{]}  & 7 \\ \hline
tomato & 18317 & {[}`can', `cup', `g', `gram', `halved', `kg', `lb', `ounce', `pound', `quart', `slice', `tablespoon'{]}  & 12 \\ \hline
black pepper & 15743 & {[}`cup', `dash', `pinch', `tablespoon', `teaspoon'{]}                            & 5  \\ \hline
milk   & 15287 & {[}`can', `cup', `g', `gallon', `gram', `liter', `ml', `ounce', `pint', `quart', `tablespoon', `teaspoon'{]}  & 12 \\ \hline
lemon juice  & 12409 & {[}`cup', `dash', `ml', `ounce', `tablespoon', `tbsp', `teaspoon', `tsp'{]} & 8 \\ \hline
pepper  & 12072 & {[}`corn', `cup', `dash', `pinch', `tablespoon', `teaspoon'{]}  & 6   \\ \hline
salt pepper  & 11019 & {[}`dash', `pinch', `teaspoon'{]}   & 3   \\ \hline
flour  & 10779 & {[}`cup', `g', `kg', `lb', `ml', `ounce', `tablespoon', `teaspoon'{]}  & 8 \\ \hline
oil   & 10651 & {[}`cup', `ml', `ounce', `quart', `tablespoon', `tbsp', `teaspoon', `tsp'{]} & 8  \\ \hline
ginger & 10519 & {[}`cm', `cup', `dash', `g', `inch', `ounce', `piece', `pinch', `slice', `tablespoon', `teaspoon', `tsp'{]}   & 12 \\ \hline
carrot  & 9981  & {[}`cup', `g', `lb', `ounce', `pound', `slice', `tablespoon'{]}   & 7 \\ \hline
parsley  & 9798  & {[}`bunch', `cup', `g', `ounce', `sprig', `tablespoon', `teaspoon'{]} & 7  \\ \hline
cinnamon   & 9257  & {[}`cup', `dash', `inch', `piece', `pinch', `stick', `tablespoon', `teaspoon'{]}  & 8  \\ \hline
\end{tabular}%
}
\caption{20 most popular ingredients, their frequency, popularly used units, and unit counts.}
\label{tab:popular_ing_unit}
\end{table*}

\begin{figure*}[!htb]
\centering
    \includegraphics[width=\textwidth, angle=0]{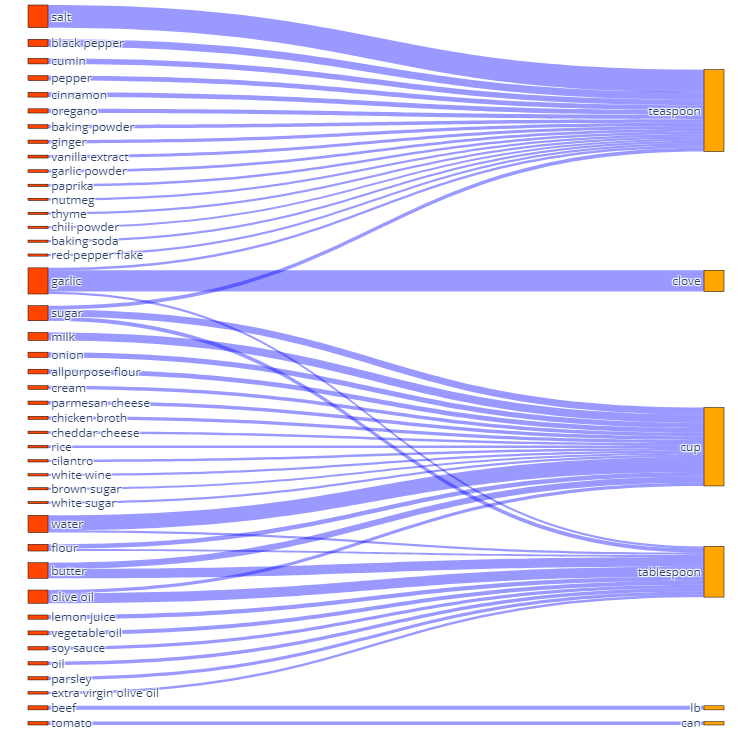}
  \caption{A list of 50 most popular ingredient-unit pairs. An ingredient may have one or more units with which it occurs in the recipe's text. For a longer list, please see Supplementary Figure S1.}
  \label{50_ing_unit_pair}
\end{figure*}

We identified 81,240 unique ingredient-unit pairs. Among these, 4,961 (6.1\%) were already in standard units (grams for solids (3,932) and milliliters for liquids (1,029)), while 93.9\% were in non-standard units. We aimed to convert all non-standard units into standard ones to ensure consistency across recipes. We manually found the standard unit conversion of each unit for each ingredient. The ingredient-unit pairs were far too many for manual conversion, hence we sorted the pairs based on the frequency and converted frequency-occurring pairs into their standard units. We successfully converted 43.90\% of the most frequently used ingredient-unit pairs into standard units. Figure~\ref{50_ing_unit_pair} illustrates 50 commonly used ingredient-unit pairs in our dataset. Ingredients such as salt, black pepper, cumin, cinnamon, oregano are widely used in teaspoons, garlic with clove, beef in lb, butter, olive oil, and vegetable oil in tablespoons, water, milk, sugar, and flour in cups.

\subsection*{Integration of RecipeDB2 to USDA ingredient}
To find the nutritional profile of ingredients, we need to accurately map ingredients from the RecipeDB2 dataset to their corresponding entries in the Standard Reference Legacy Release database~(\href{https://fdc.nal.usda.gov/}{https://fdc.nal.usda.gov/}) (United States Department of Agriculture; USDA). RecipeDB2 presents 128,942 recipes comprising 35,474 ingredients from 32 regions, while USDA has 1,050 unique ingredients. The disparity in the number of entities between these databases is due to the many-to-one mappings of RecipeDB ingredients to that in USDA. A typical USDA ingredient occurs in recipes in various avatars.

We implemented three mapping strategies: Jaccard similarity~\cite{bag2019efficient}, BERT-based~\cite{kenton2019bert, reimers2019sentence}, and RoBERTa-based~\cite{liu2019roberta}. Each approach leverages different techniques for identifying the best matches between ingredients based on textual similarity and contextual understanding. The Jaccard similarity is a well-known metric that compares two sets by measuring the intersection over the union. We compared the ingredient names from RecipeDB2 and USDA by breaking them down into sets of tokens (e.g., ``ground cumin'' becomes {ground, cumin}). The Jaccard score was calculated by dividing the number of shared words (intersection) by the total number of unique words (union) between two ingredients. BERT (Bidirectional Encoder Representations from Transformers) is a language model that captures the contextual meaning of words and phrases. We used a pre-trained BERT model for ingredient mapping to encode the RecipeDB2 and USDA ingredient names into dense vector representations. These vectors were then compared using cosine similarity to find the closest match. RoBERTa (Robustly Optimized BERT Pretraining Approach) is an improved variant of BERT designed with more robust training strategies and adjustments. Similar to BERT, RoBERTa encodes text into contextualized vector representations, which we used to compute the similarity between RecipeDB2 and USDA ingredient names.

\begin{figure}[!htb]
\centering
    \includegraphics[width=\textwidth, angle=0]{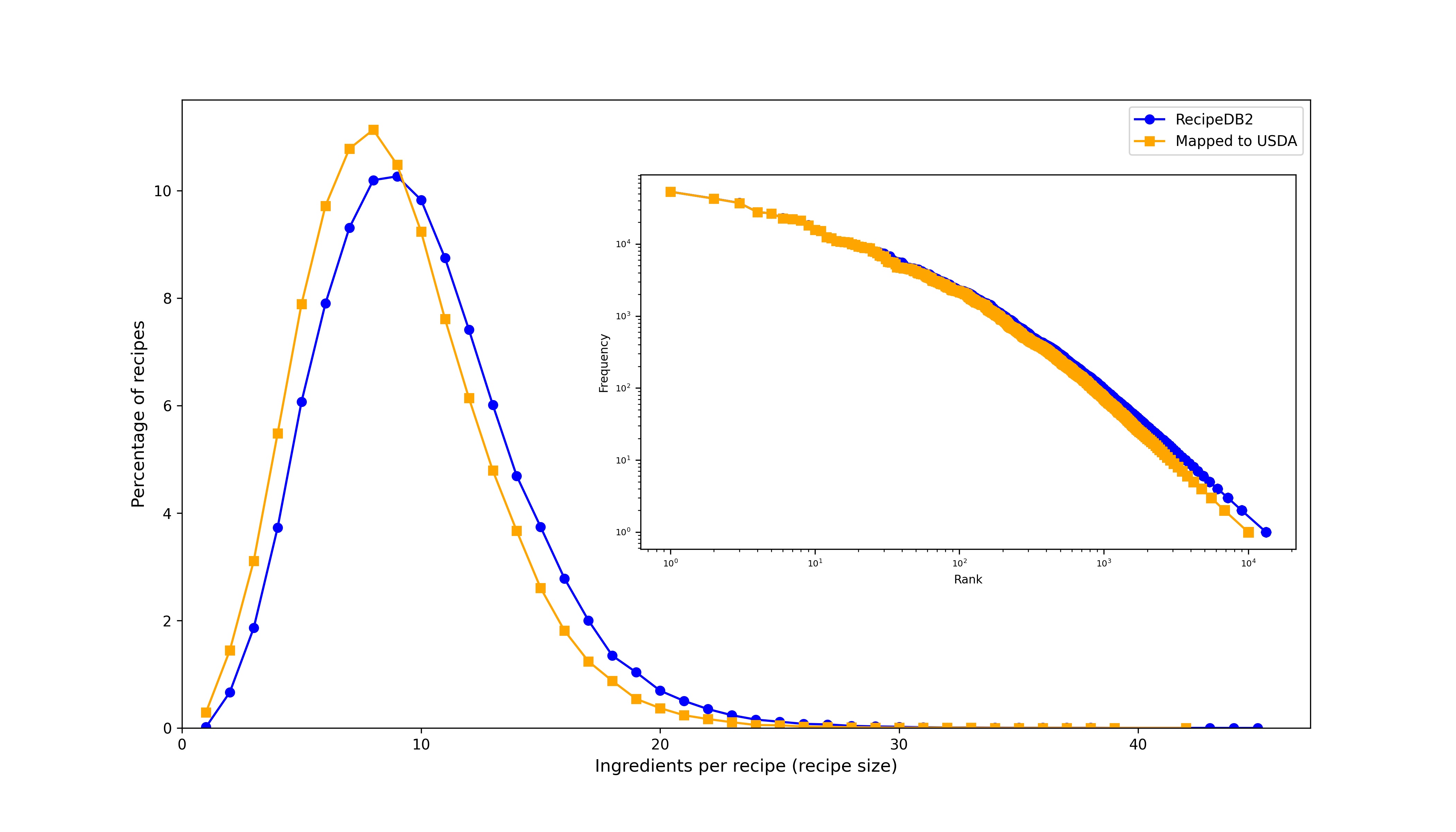}
  \caption{Recipe size distribution before and after mapping to USDA suggesting no significant difference between the two. (inset) Frequency rank distribution before and after mapping to USDA.}
  \label{rec_size}
\end{figure}

We manually evaluated the mapping results for the 200 most popular ingredients. Table~\ref{tab:model_perf} shows the detailed performance comparison of RecipeDB2 ingredient mapping to that of USDA. The BERT model outperformed the RoBERTa and Jaccard models, achieving the best overall performance with F1 scores and accuracy of 87.90\% and 79.50\%, respectively, making it the preferred method for ingredient-to-USDA mapping. Out of 35,474 ingredients in RecipeDB2, 25,903 were successfully mapped to USDA using BERT with a similarity score of $\geq$ 70\%. Figure~\ref{rec_size} (inset) shows the most frequently used ingredients mapping using the BERT strategy. This process allows us to retrieve nutritional information for approximately 73.01\% of the ingredients, which we then use to calculate the nutritional profiles of the recipes.


\begin{table*}[]
\centering
\begin{tabular}{|c|c|c|c|c|c|}
\hline
Top X Ingredients    & Model    & Accuracy & Precision & Recall & F1 Score \\ \hline
\multirow{3}{*}{10}  & \textbf{BERT}   & \textbf{80.00}  & \textbf{80.00}  & \textbf{100} & \textbf{88.88}    \\ \cline{2-6} 
                     & RoBERTa  &  60.00  & 60.00  &  100   &  75.00  \\ \cline{2-6} 
                     & Jaccard & 20.00  & 40.00  & 28.57 & 33.33    \\ \hline
\multirow{3}{*}{20}  & \textbf{BERT}  &  \textbf{70.00} & \textbf{70.00}  & \textbf{100}  & \textbf{82.35}    \\ \cline{2-6} 
                     & RoBERTa  & 50.00   & 50.00  & 100  &  66.67  \\ \cline{2-6} 
                     & Jaccard & 25.00  & 57.14 & 25.00  & 34.78    \\ \hline
\multirow{3}{*}{50}  & \textbf{BERT} & \textbf{82.00}  & \textbf{80.43} & \textbf{100}  &  \textbf{89.15}    \\ \cline{2-6} 
                     & RoBERTa  &  52.00   & 51.02  &  100 &   67.57  \\ \cline{2-6} 
                     & Jaccard & 32.00  & 76.47       &  30.23 &  43.33  \\ \hline
\multirow{3}{*}{100} & \textbf{BERT} & \textbf{84.00}  & \textbf{82.79} & \textbf{100}  & \textbf{90.58}    \\ \cline{2-6} 
                     & RoBERTa  &  60.00  &  58.76  & 100 &  74.02   \\ \cline{2-6} 
                     & Jaccard & 30.00  & 86.20 & 27.47     & 41.66    \\ \hline
\multirow{3}{*}{200} & \textbf{BERT}  & \textbf{79.50} & \textbf{81.42}     & \textbf{95.51}  & \textbf{87.90}    \\ \cline{2-6} 
                     & RoBERTa  & 56.50  & 56.02   & 97.27 &  71.09   \\ \cline{2-6} 
                     & Jaccard & 25.50   & 97.72       & 23.11  & 37.38    \\ \hline
\end{tabular}
\caption{Comparison of performance of models in accurately mapping RecipeDB2 to USDA ingredient using metrics accuracy, precision, recall, and F1 score. The word embeddings were generated for the BERT, RoBERTa, and Jaccard models.}
\label{tab:model_perf}
\end{table*}

\subsection*{Nutritional profile of recipes}
The nutritional profile of a recipe was calculated by aggregating the nutritional values of each ingredient, factoring in their respective quantities, and applying standard unit conversions. By mapping ingredient measurements to standardized values, this approach reasonably estimates various nutritional components, including calorific content, macro-nutrients (such as proteins, fats, and carbohydrates), and essential micro-nutrients (like vitamins and minerals). This comprehensive analysis offers valuable insights into the nutritional composition of recipes, enabling a deeper understanding of how different ingredients contribute to the overall nutritional profile of a dish. It also facilitates comparisons across recipes, dietary trends, and regional culinary practices, enabling one to make informed choices about nutrition and meal planning.

RecipeDB2 presents a repository of 128,942 recipes. After applying the ingredient mapping strategy, we successfully mapped ingredients to 128,899 recipes (fully or partially), demonstrating that our approach covers the majority of the dataset. However, 43 recipes remained completely unmapped due to the missing ingredient data. These entirely unmapped recipes have fewer ingredients, with a minimum of 1 to a maximum of 8 (Christmas Truffle) (see Supplementary Figure S2). On average, each unmapped recipe contained 2.5 ingredients, for a total of 115 unmapped ingredients across all 43 recipes. This indicates that our mapping strategy efficiently maps the most popular ingredients and recipes with a larger number of ingredients. The recipe size distribution (in Figure~\ref{rec_size} indicates that before mapping, the average recipe size was 9.978, which marginally reduced to 8.943 post-mapping, suggesting that our approach successfully maps most ingredients to their USDA nutritional profile.


The macro-nutrient analysis of the RecipeDB2 dataset provides insightful statistics that reflect the diversity and composition of recipes within the dataset (see Supplementary Figure S3). The protein content across recipes is from 0 to 285.84 grams, indicating that some recipes may not contain protein sources and some recipes with protein-rich ingredients such as legumes, meat, and dairy. The average protein content across all recipes is 49.48 grams, indicating a moderate protein level in most recipes. Carbohydrate content significantly varies from 0 to 1203.84 grams, pointing to recipes such as baked goods or pasta dishes containing carbohydrate-dense ingredients. The average carbohydrate content is 245.20 grams, reflecting a balanced proportion of carbohydrates in the recipes~\cite{trattner2017estimating}. Total lipid content in the dataset ranges from 0 to 660.41 grams, indicating the influence of fat content in recipes is likely influenced by the use of oils, nuts, and fatty cuts of meat. The average lipid content is 102.02 grams. The energy values contain considerable diversity, ranging from 0 to 8745 calories, with an average energy of 1989.01 calories across the recipes.

\subsection*{Predicting the category of ingredients}
We manually categorized 10,659 ingredients out of a total of 35,474 from RecipeDB~\cite{batra2020recipedb}, which is approximately 30\% of the dataset. Left with a large number of ingredients with no category label, we implemented machine learning models to predict the category for the remaining 24,815 ingredients (70\%). Analysis of the uncategorized ingredients revealed a wide distribution in frequency, with some ingredients appearing only once and some appearing in 8,844 recipes. On average, each uncategorized ingredient appeared approximately 5.5 times. This suggests that one should prioritize manually labeling the categories of most frequently used ingredients.

We represented each ingredient name as TF--IDF n-gram vectors. Models were trained and evaluated on the 10,659 manually labelled ingredients using a stratified 80/20 train--test split with stratification to preserve the class distribution across folds. The models aim to automate the classification process and improve efficiency in assigning category to each ingredient. We compared seven supervised classifiers: Random Forest (RF), Decision Tree (DT), Support Vector Machine (SVM), Logistic Regression (LR), Gradient Boosting (GB), Naive Bayes (NB), and K-Nearest Neighbors (KNN). Performance is reported in Table~\ref{tab:cat_performance} as weighted-averaged accuracy, precision, recall, and F1 score across all 34 categories. Random Forest achieved the best performance (accuracy 89.12\%, weighted F1 89.16\%), with the tree-ensemble models (RF, DT) and the margin-based SVM clearly outperforming the linear (LR, GB) and distance/probability-based (NB, KNN) baselines. This ordering is consistent with the high-dimensional, sparse, and lexically redundant nature of ingredient-name features, which tree ensembles partition effectively while distance- and density-based methods (KNN, NB) degrade. 

RecipeDB2 organizes ingredients into 34 categories, the largest being Dish (3,723), Spice (3,314), Vegetable (3,303), Meat (3,261), Dairy (2,437), and Additive (2,313) (Supplementary Figure S4); Supplementary Figure S5 shows the regional composition of these categories. The predicted categories feed directly into the dietary-style annotation rules and the category-level nutritional analysis, so the classifier's reliability on rare categories bounds the accuracy of those downstream layers.

\begin{table}[]
\centering
\begin{tabular}{|c|c|c|c|c|}
\hline
Model  & Accuracy & Precision & Recall   & F1 Score \\ \hline
\textbf{RF} & \textbf{89.12} & \textbf{89.38}  & \textbf{89.11} & \textbf{89.16} \\ \hline
DT   & 87.19 & 87.44  & 87.18 & 87.25 \\ \hline
SVM & 85.31 & 86.06 & 85.31 & 85.48 \\ \hline
LR  & 82.04 & 83.17  & 82.04 & 82.16 \\ \hline
GB & 80.74 & 83.05  & 80.74 & 81.09 \\ \hline
NB & 70.20 & 71.86  & 70.19 & 67.72 \\ \hline
KNN & 67.78 & 71.83  & 67.79 & 68.92 \\ \hline
\end{tabular}
\caption{Performance of machine learning models in predicting the category of each ingredient in terms of accuracy, precision, recall, and F1 score.}
\label{tab:cat_performance}
\end{table}

\subsection*{Dietary style annotations}
We annotated each recipe with a dietary style using a deterministic rule set over the predicted ingredient categories. The rules test for the presence or absence of five animal-derived category groups, such as meat, eggs, dairy, fish, and seafood, together with a \textit{composite-dish} guard described below. We adopt an exclusion-based (conservative) policy: a recipe receives a vegetarian/vegan label only when no ingredient from a disallowed group is present, so that mislabelling errs toward under-claiming rather than over-claiming a restrictive diet. A \textit{composite dish} is an ingredient whose entry names a prepared or multi-component item rather than a single raw ingredient, and whose internal composition is therefore opaque to category-level rules. Because such items may conceal animal-derived components (e.g., anchovy in Worcestershire sauce, animal gelatin), any recipe containing an unresolved composite dish is excluded from the positive vegan/vegetarian classes.

\paragraph{Labelling rules.}
Let a recipe's ingredient set be summarized by the indicator variables $\mathrm{meat}$, $\mathrm{egg}$, $\mathrm{dairy}$, $\mathrm{fish}$, $\mathrm{seafood}$, and $\mathrm{composite}$. A recipe is labelled:
\begin{itemize}
  \item \textbf{Vegan} -- if none of meat, egg, dairy, fish, seafood, or composite are present;
  \item \textbf{Lacto-Vegetarian} -- if dairy is present, and meat, egg, fish, seafood, and composite are absent;
  \item \textbf{Ovo-Vegetarian} -- if egg is present, and meat, dairy, fish, seafood, and composite are absent;
  \item \textbf{Ovo-Lacto-Vegetarian} -- if egg and dairy are present, and meat, fish, seafood, and composite are absent;
  \item \textbf{Pescetarian} -- if fish or seafood is present, and meat and composite are absent.
\end{itemize}
The rules are applied in this order, and by construction every recipe receives exactly one label, making the assignment mutually exclusive and exhaustive over the dataset.

\section*{Webserver implementation}
RecipeDB2 facilitates exploring and understanding the intricate relationships between cuisines, recipes, and ingredients. With interactive data visualizations and a range of intuitive search options, users can quickly access relevant information, uncover culinary patterns, and gain deeper insights. The platform supports in-depth analysis of recipes and ingredients, helping users explore the rich diversity of global cuisines.

Built as a relational database using MongoDB, RecipeDB2 is optimized for efficient data management and querying. The web server is powered by Express, a NodeJS-based web development framework, renowned for its minimalistic and unopinionated design. Express facilitates seamless integration with MongoDB using libraries like Mongoose for schema modeling and efficient querying. The frontend of RecipeDB2 is developed using ReactJS, a framework for building frontend, ensuring a responsive and user-friendly interface. The platform employs an NGINX HTTP server to enhance further performance, routes requests to the Express application, and enables data compression, leading to faster page load times. ElasticSearch engine would significantly improve the platform's search capabilities by allowing for faster, full-text searches and more advanced query features, particularly in handling large datasets.

RecipeDB2~(\href{https://cosylab.iiitd.edu.in/recipedb2/}{https://cosylab.iiitd.edu.in/recipedb2/}) is optimized for modern web browsers, providing the best experience on the latest versions of Google Chrome, Firefox, Opera, Internet Explorer, and Microsoft Edge. This cross-browser compatibility ensures smooth navigation and interaction across different platforms and devices.

\section*{Use cases}
\subsection*{Searching recipes by cuisine}
RecipeDB2 enables users to search for recipes by cuisine at the `region' or `country' level. For example, users can explore the Mediterranean region and Italian as a country, or search for Italian cuisine directly. Each search field provides an autosuggest function. The results page displays a comprehensive list of Italian recipes, including recipe names, estimated macro-nutrient information, and links to individual recipe pages. Clicking on a recipe name directs users to a detailed recipe page, while the `More Info' tab provides a pop-up with an extensive nutritional profile, covering both macro- and micro-nutrients as provided by the USDA. 

\subsection*{Searching recipe using macro-nutrients}
RecipeDB2 allows users to search for recipes based on their nutritional profiles, focusing on macro-nutrients such as fats, protein, carbohydrates, and energy. This feature allows users to identify recipes that meet specific dietary requirements or preferences. Additionally, a detailed breakdown of the nutritional profile is available, offering insights into the ingredients that contribute to the overall nutritional value of each recipe. Such nutritional transparency helps users make informed choices about their meals, catering to their health and dietary goals.

\subsection*{Multi-search and Multi-attribute search for ingredients and categories}
The implementation of multi-search and multi-attribute search capabilities for ingredients and categories significantly enhances the user experience by offering flexibility in refining recipe results. Users can conduct simultaneous searches using both ingredients included and excluded, allowing for more nuanced filtering that caters to individual dietary preferences and restrictions. The advanced search feature further empowers users by enabling them to apply multiple criteria within the `Ingredients Used' and `Ingredients Not Used' tabs, ensuring a better search experience. Additionally, users can filter recipes based on both used and unused categories simultaneously, with a choice to select multiple values within each category.

\section*{Discussion}
RecipeDB2 is a comprehensive database that enhances culinary knowledge. RecipeDB2 offers a detailed recipe dataset, capturing information on ingredients, preparation methods, cooking techniques, cultural origins, and nutritional profiles. The user-friendly RecipeDB2 interface enriches the user experience, facilitating a better exploration and understanding of recipes across diverse culinary contexts. By integrating the culinary context of the recipes and ingredients along with their nutritional correlates, RecipeDB2 empowers users to explore culinary relationships. 

One of the primary directions in which such a structured dataset can be driven is that of generating novel recipes by applying large language models~\cite{goel2022ratatouille}. In principle, such a novel recipe generation engine can be tailored to individual user preferences, dietary restrictions, ingredient availability, cost, and calorie content using the RecipeDB2 dataset. The integration of culinary and nutritional details of recipes enables the users to make informed dietary choices and promotes healthier eating habits. The database serves as a valuable resource for exploring the cultural and historical significance of various cuisines and ingredients. 

When compared to other existing recipe datasets, such as Recipe1M~\cite{marin2021recipe1m+}, Epicurious, Yummly, RecipeNLG~\cite{bien2020recipenlg}, and RecipeDB~\cite{batra2020recipedb}, RecipeDB2 stands out for its structured and comprehensive approach to culinary data. RecipeDB2 marks a huge step forward in culinary data by identifying named entities, mapping ingredients to find the nutritional profile, and predicting the category of ingredients. By incorporating metadata such as preparation methods, cultural origins, dietary style, and nutritional information, RecipeDB2 improves data richness and accessibility. This systematic approach enables more nuanced analysis and applications in culinary research while effectively resolving the limits of unorganized datasets. RecipeDB2 empowers researchers, chefs, and food enthusiasts to delve deeper into the complex world of food. This comprehensive approach broadens culinary expertise and emphasizes on the significant influence of food on our lives and cultures. Future work will focus on expanding the database's capabilities by integrating domain ontologies such as FoodOn and LanguaL to improve ingredient normalization and enable semantic reasoning across recipes, enhancing user interactivity, and promoting research in culinary sciences. We also aim to incorporate food-chemical relationships using external resources such as FooDB and FlavorDB2, facilitating detailed flavor and health-centric exploration. On the analytical front, we plan to introduce machine-learning pipelines for nutrient prediction~\cite{ispirova2024msgen}, ingredient substitution, and recipe rating models.

\section*{Authors' contribution}
M.G. scraped and cleaned the dataset, conducted the experiments, and analyzed the results. M.G., S.B., and S.S. developed the webserver. M.G., M.P., H.P., and S.V.M. manually converted the data into standard units. G.B. supervised the project. M.G. and G.B. wrote and reviewed the manuscript.

\section*{Competing interests}
No competing interest is declared.

\section*{Data Availability Statement}
The datasets generated and analyzed during the current study are not publicly available due to institutional copyright restrictions.

\section*{Acknowledgments}
GB thanks Indraprastha Institute of Information Technology Delhi (IIIT-Delhi) for the computational support. GB thanks Technology Innovation Hub (TiH) Anubhuti for the research grant. MG is a research scholar in the Complex Systems Laboratory and is thankful to IIIT-Delhi for the research fellowship. SB, SS, MP, and HP are research interns and are thankful to IIIT-Delhi. This study was supported by the Infosys Centre for Artificial Intelligence and Center of Excellence in Healthcare, IIIT-Delhi.

\section*{Supplementary data}
Supplementary Tables and Figures are provided in the Supplementary Information at the end of this document.


\clearpage
\setcounter{figure}{0}
\setcounter{table}{0}
\renewcommand{\thefigure}{S\arabic{figure}}
\renewcommand{\thetable}{S\arabic{table}}

\section*{Supplementary Information}

\noindent Supplementary Table~\ref{tab:continent_region} shows the mapping of RecipeDB2 continent and region, along with the recipe count.

\begin{table*}[!htb]
\centering
\begin{tabular}{|c|c|c|}
\hline
Continent      & Region                 & Recipe Count \\ \hline
African        & Middle Eastern         & 325          \\ \hline
African        & Northern Africa        & 1610         \\ \hline
African        & Rest Africa            & 2774         \\ \hline
Asian          & Chinese                & 5954         \\ \hline
Asian          & Indian Subcontinent    & 13780        \\ \hline
Asian          & Japanese               & 2063         \\ \hline
Asian          & Korean                 & 673          \\ \hline
Asian          & Middle Eastern         & 3674         \\ \hline
Asian          & Mongolian              & 75           \\ \hline
Asian          & Southeast Asian        & 2105         \\ \hline
Asian          & Thai                   & 2724         \\ \hline
Australasian   & Australian             & 5819         \\ \hline
European       & Belgian                & 1060         \\ \hline
European       & Continental            & 1784         \\ \hline
European       & Deutschland            & 4321         \\ \hline
European       & Eastern European       & 2504         \\ \hline
European       & French                 & 6485         \\ \hline
European       & Greek                  & 4208         \\ \hline
European       & Irish                  & 2529         \\ \hline
European       & Italian                & 16983        \\ \hline
European       & Middle Eastern         & 1            \\ \hline
European       & Portuguese             & 687          \\ \hline
European       & Scandinavian           & 2811         \\ \hline
European       & Spanish                & 2163         \\ \hline
European       & Spanish and Portuguese & 70           \\ \hline
European       & UK                     & 4430         \\ \hline
European       & Western European       & 42           \\ \hline
Latin American & Caribbean              & 3030         \\ \hline
Latin American & Central American       & 455          \\ \hline
Latin American & Mexican                & 14656        \\ \hline
Latin American & South American         & 7171         \\ \hline
North American & Canadian               & 6694         \\ \hline
North American & US                     & 5126         \\ \hline
Unknown        & Unknown                & 72           \\ \hline
\end{tabular}
\caption{Mapping of cuisines from their (32) Regions to (7) Continents and (region-wise) number of recipes.}
\label{tab:continent_region}
\end{table*}

\clearpage
\begin{figure*}[!htb]
    \centering
    \includegraphics[width=\textwidth, angle=0]{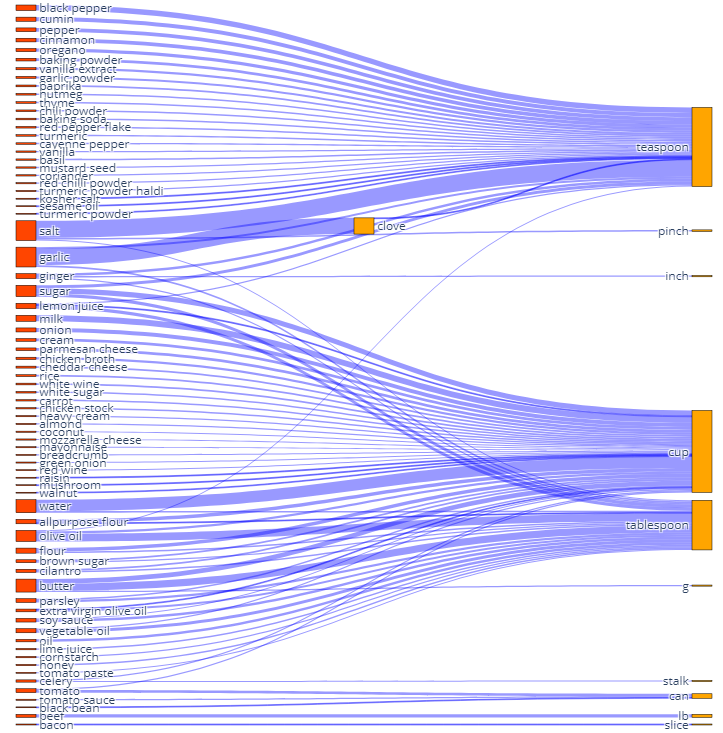}
    \caption{Visualization of the 100 most popular ingredient-unit pairs derived from the dataset. This figure illustrates the frequency of each ingredient along with its corresponding measurement unit, providing insights into common culinary practices. For example, ingredients such as `sugar' are frequently measured in cups, while `spices' are often measured in teaspoons or grams.}
    \label{fig:100_ing_unit_pair}
\end{figure*}

\clearpage
\begin{figure*}[!htb]
    \centering
    \includegraphics[width=\textwidth, angle=0]{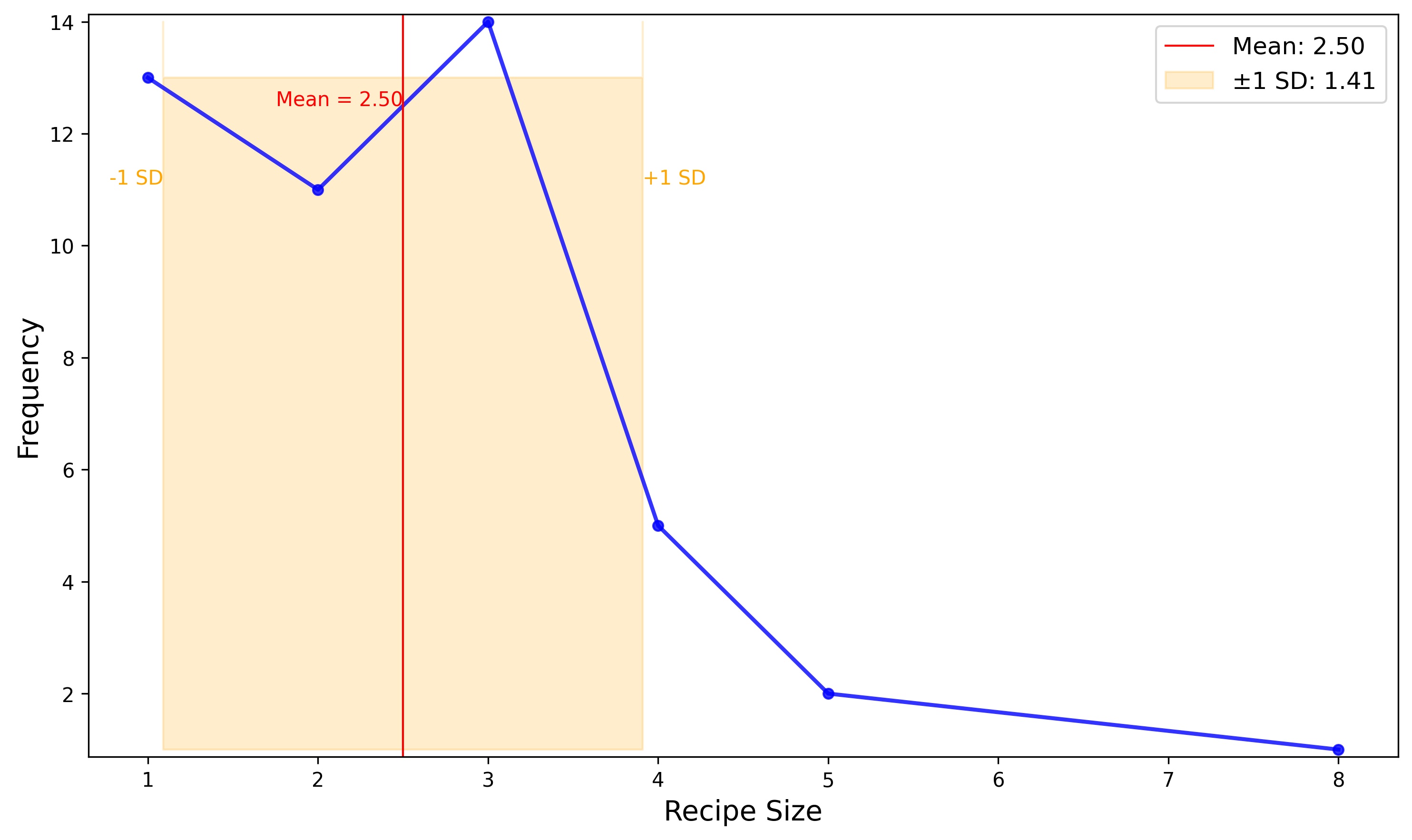}
    \caption{Statistics of 43 recipes that were left out of the BERT mapping protocol. The distribution of the recipe sizes has an average of 2.5, indicating that most of the recipes for which no ingredient was mapped to USDA consisted of a small number of ingredients. Christmas Truffle was an outlier, exhibiting a significantly larger recipe size of 8, suggesting it may involve ingredients that are difficult to find a corresponding USDA match.}
    \label{fig:supplementary_figure_1}
\end{figure*}

\clearpage
\begin{figure*}[!htb]
    \centering
    \includegraphics[width=\textwidth, angle=0]{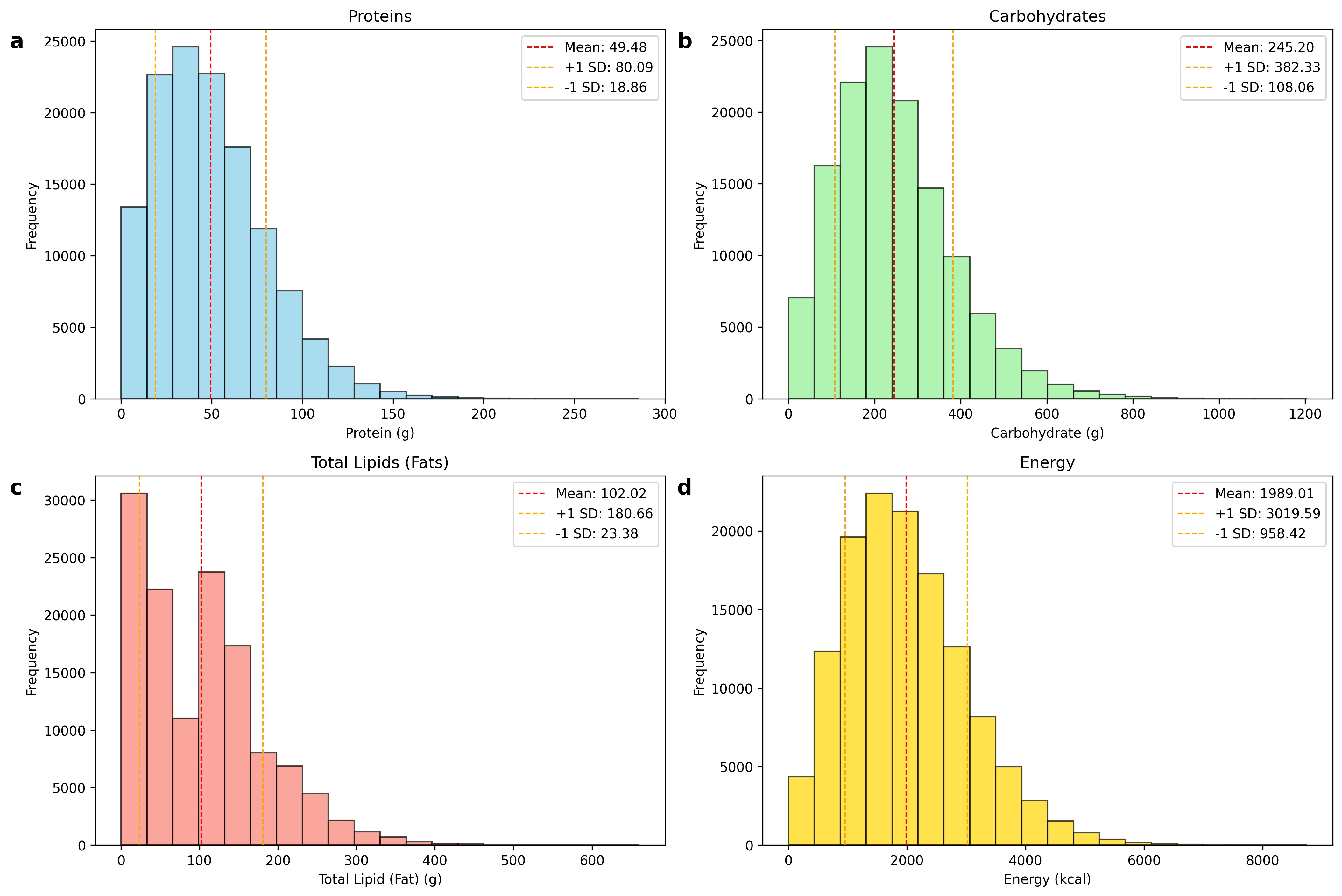}
    \caption{Macro-nutrient analysis of recipes illustrating the distribution of key nutritional components. a. Protein distribution (grams), b. Carbohydrates (grams), c. Total lipids (fats) (grams), and d. Energy (calories). The data highlights variations in macro-nutrient composition across the recipes along with their mean and standard deviation, thus providing insights into the scatter of these macro-nutrients.}
    \label{macro_analysis}
\end{figure*}

\clearpage
\begin{figure*}[!htb]
    \centering
    \includegraphics[width=\textwidth, angle=0]{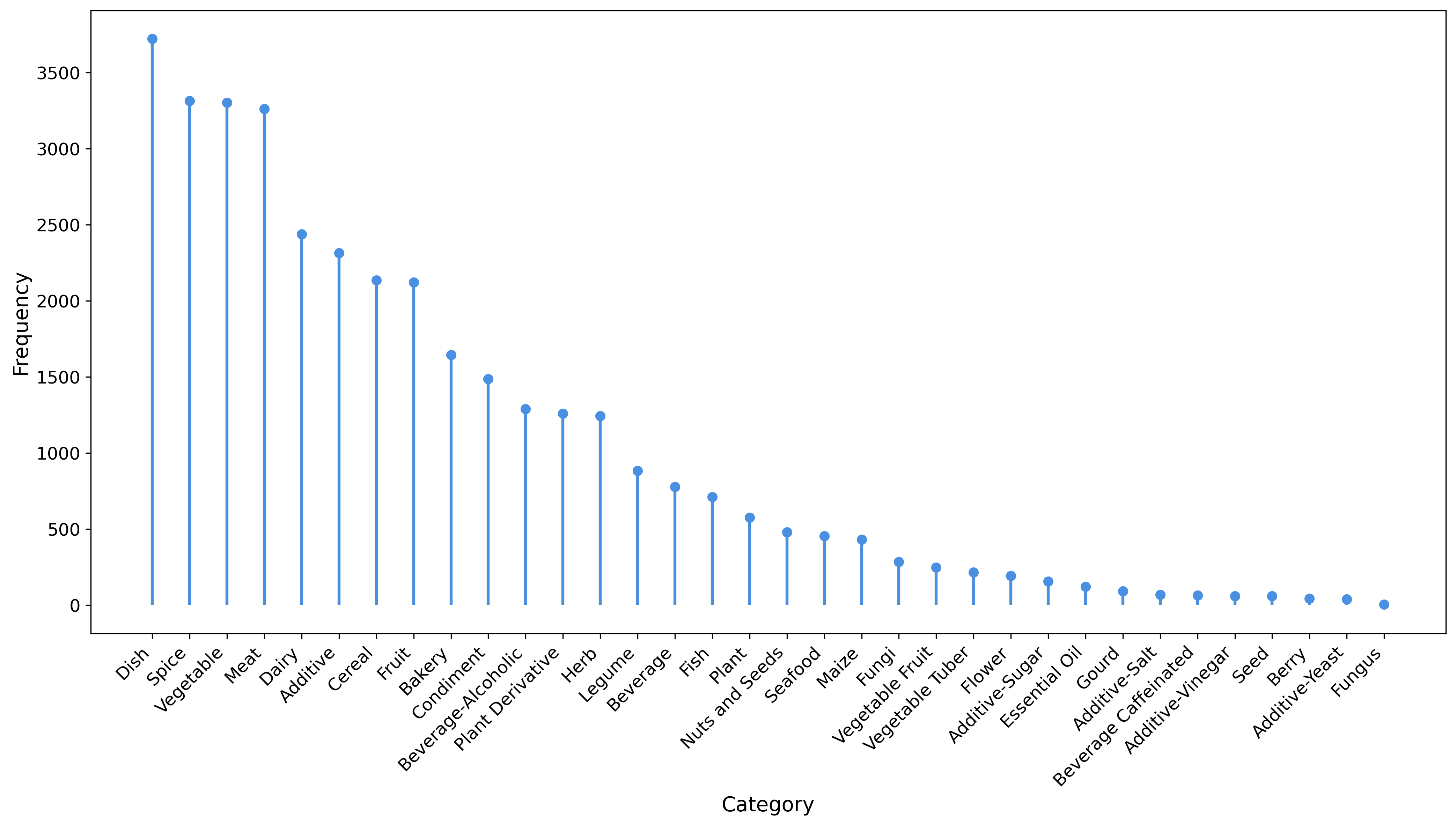}
    \caption{Frequency of ingredients in each of the 34 categories after implementing the Random Forest-based strategy. A total of 35,474 ingredients were mapped, of which 10,659 ($\sim$30\%) most frequently occurring were mapped manually, and the rest 24,815 ($\sim$70\%) were done using the automated, machine learning-based protocol.}
    \label{ing_cat}
\end{figure*}

\clearpage
\begin{figure*}[!htb]
    \centering
    \includegraphics[width=\textwidth, angle=0]{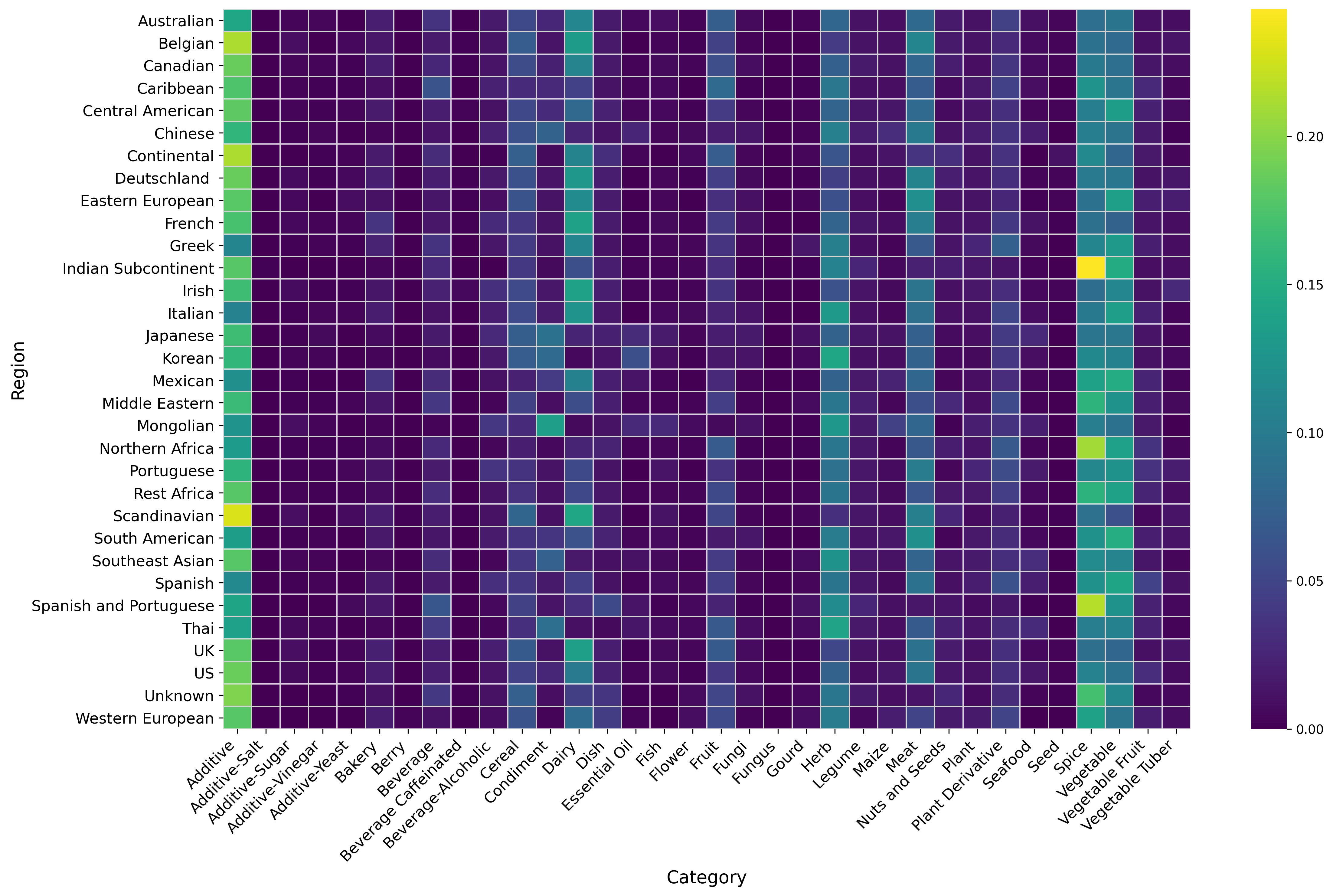}
    \caption{Ingredient category composition of recipes for different cuisines (Region). For each cuisine, the heatmap represents the fraction of ingredients belonging to each ingredient category, highlighting the dominant and less prevalent ingredient categories in each cuisine. Additives, Spices, and Vegetables are among the most dominant ingredient categories, with a very heavy representation in the recipes, in general.}
    \label{ing_heatmap}
\end{figure*}

\end{document}